\documentclass[%
 reprint,
 amsmath,amssymb,
 aps,
]{revtex4-1}

\usepackage{amsthm}
\usepackage{xcolor}
\usepackage{graphicx}
\usepackage{bm}
\usepackage{mathrsfs}
\usepackage{hyperref}
\usepackage[normalem]{ulem}
\usepackage{lipsum}
\usepackage{IEEEtrantools}
\usepackage[version=3]{mhchem}
\usepackage{verbatim}
\usepackage{siunitx}
\usepackage{enumitem}
\usepackage{textcomp}
\usepackage{booktabs}
\usepackage{microtype}
\bibpunct{[}{]}{;}{n}{}{}

\setlist[itemize]{noitemsep, topsep=0pt}
\usepackage[margin=1in]{geometry}
\usepackage{amsmath}
\usepackage{pifont}

\theoremstyle{definition}
\newtheorem{assumption}{Assumption}

\theoremstyle{plain}
\newtheorem{lemma}{Lemma}
\newtheorem{theorem}{Theorem}[section]

\newcommand{\cmark}{\ding{51}} 
\newcommand{\xmark}{\ding{55}} 

\renewcommand{\toprule}{\specialrule{1.pt}{2pt}{2pt}}
\renewcommand{\midrule}{\specialrule{0.6pt}{2pt}{2pt}}
\renewcommand{\bottomrule}{\specialrule{0.8pt}{2pt}{2pt}}

\begin{document}

\title{
Latent Point Collapse  \\
on a Low Dimensional Embedding \\
in Deep Neural Network Classifiers}

\author{Luigi Sbailò$^{1,2*}$, and Luca M. Ghiringhelli$^{1,2,3}$}
\affiliation{
$^1$Physics Department and IRIS Adlershof of the Humboldt-Universität zu Berlin, Berlin, Germany.\\
$^2$Department of Materials Science and Engineering, Friedrich-Alexander Universität, Erlangen-Nürnberg, Germany.\\ 
$^3$Erlangen National High Performance Computing Center (NHR@FAU), Friedrich-Alexander Universität  Erlangen-Nürnberg, Germany
$^*$email:  luigi.sbailo@gmail.com; 
}
\date{\today}

\begin{abstract}
The configuration of latent representations plays a critical role in determining the performance of deep neural network classifiers. In particular, the emergence of well-separated class embeddings in the latent space has been shown to improve both generalization and robustness. In this paper, we propose a method to induce the collapse of latent representations belonging to the same class into a single point, which enhances class separability in the latent space while enforcing Lipschitz continuity in the network.
We demonstrate that this phenomenon, which we call \textit{latent point collapse}, is achieved by adding a strong $L_2$ penalty on the penultimate-layer representations and is the result of a push-pull tension developed with the cross-entropy loss function.
In addition, we show the practical utility of applying this compressing loss term to the latent representations of a low-dimensional linear penultimate layer.
The proposed approach is straightforward to implement and yields substantial improvements in discriminative feature embeddings, along with remarkable gains in robustness to input perturbations.
\end{abstract}

\maketitle

\section{Introduction}

Deep neural networks (DNNs) excel in various tasks, but they often struggle with ensuring robust performance and reliable generalization.
A key insight into addressing these challenges lies in understanding and controlling the geometry of the latent representations that DNNs learn. 
In particular, it has been observed that increasing the margin between classes, i.e., making classes more separable in the latent space, can yield significant gains in both robustness and generalization \cite{elsayed2018largemargindeepnetworks, wei2021improvedsamplecomplexitiesdeep, ding2020mmatrainingdirectinput, Sokolic_2017}.
Indeed, the relationship between generalization and robustness is well-established in the literature \cite{xu2010robustnessgeneralization, szegedy2014intriguing, achille2018emergenceinvariancedisentanglementdeep, novak2018sensitivitygeneralizationneuralnetworks, zhang2019theoreticallyprincipledtradeoffrobustness}. 

DNNs naturally tend to improve the separation of different classes in the latent space during training, and this process occurs at a constant geometric rate \cite{he2023law}. 
Such evolving separation manifests in the phenomenon of \textit{neural collapse} (NC) \cite{neural_collapse, han2022neural}, which emerges in the penultimate layer of DNN classifiers, particularly in overparameterized models during the terminal phase of training (TPT). 
Even after the point of zero training error, the network further refines the representations by increasing their relative distances in the latent space. In practice, this means that the class means in the penultimate layer collapse to the vertices of an equiangular tight-frame simplex (ETFS).
The occurrence of NC has been observed in large language models \cite{wu2024linguisticcollapseneuralcollapse} and extensively investigated in theoretical studies of unconstrained feature models \cite{unconstrained2020, ErgenPilanci2020, NEURIPS2021_geometricAnalysis, dissectingContrastingLearning, ji2022unconstrained, tirer2022extended, zhou2022optimization, fisher2024pushingboundariesmixupsinfluence} and layer-peeled models \cite{Fang_2021, liu2023generalizingdecouplingneuralcollapse}. Nonetheless, perfect convergence to an ETFS is not always observed in practical scenarios \cite{tirer2023perturbation}.

NC implies convergence to a neural geometry that enhances separability in the latent space, ultimately improving generalization and robustness.
In fact, it has been shown that these metrics continue to improve during the TPT \cite{neural_collapse}, precisely when the latent representations approach an ETFS.
Subsequent research has revealed that NC also brings other benefits. For instance, in \cite{NC_transfer_learning, galanti2022improved, li2024understandingimprovingtransferlearning, munn2024impactgeometriccomplexityneural}, NC has been linked to improved transfer learning. Building on this connection, \cite{Wang_2023_ICCV, yang2023neural} use NC-based metrics to enhance the transferability of models.
Another direction relates NC with out-of-distribution (OOD) detection: \cite{haas2023linking} show that the appearance of NC facilitates OOD detection, and \cite{ammar2023neco} propose a method leveraging NC geometry to enhance OOD detection.

The connection between margin-based approaches and robust generalization can also be understood through the lens of the Information Bottleneck (IB) principle \cite{tishby2000informationbottleneckmethod, tishby2015deep}. The IB principle suggests that DNNs seek compact yet sufficiently informative latent representations by minimizing the mutual information between inputs and latent representations, while preserving information relevant for prediction.
Empirically, it has been shown that IB improves network performance \cite{survey}, and theoretical work provides rigorous arguments for IB’s role in controlling generalization errors \cite{kawaguchi2023doesinformationbottleneckhelp}.
In practice, DNN training reveals two distinct phases: an empirical risk minimization phase, where the network primarily fits the data, followed by a compression phase, where the network constructs more compact embeddings layer by layer \cite{shwartzziv2017openingblackboxdeep}. 
This compression aligns with margin maximization and NC, suggesting that the quest for efficient representations manifests in both information-theoretic and geometric properties.

A separate line of research for improving robustness in DNNs focuses on developing Lipschitz continuous networks, as Lipschitz constraints help ensure bounded responses to input perturbations \cite{zhang2022rethinkinglipschitzneuralnetworks,cisse2017parsevalnetworksimprovingrobustness,tsuzuku2018lipschitzmargintrainingscalablecertification,Pauli_2022}. Specifically, in Lipschitz networks, the smallest perturbation required for misclassification is proportional to the Lipschitz constant \cite{zhang2022rethinkinglipschitzneuralnetworks, li2019preventinggradientattenuationlipschitz}.

\subsection{Contributions}
In this paper, we present a method to induce the collapse of all latent representations in the penultimate layer of each class into a single distinct point. 
We refer to this phenomenon as \textit{latent point collapse} (LPC). It is achieved through the imposition of a strong $L_2$ penalty on the penultimate-layer latent representations in a DNN classifier that also minimizes cross-entropy. 
We show that this strong $L_2$ penalty creates a push-pull tension with cross-entropy, ultimately leading to LPC. 
We additionally demonstrate the practical utility of applying this compressing loss to a low-dimensional linear penultimate layer. More specifically, our proposed approach:
\begin{enumerate}
    \item Uses a linear penultimate layer.
    \item Restricts its dimensionality to be low.
    \item Imposes a strong $L_2$ loss on its latent representations.
\end{enumerate}

While LPC is conceptually related to NC, there is a crucial difference: NC does not require that all same-class latent representations collapse to a single point. In LPC, however, all points from a given class converge to the same location at a fixed distance from the origin. This makes the network Lipschitz continuous and provides stronger robustness guarantees than standard training. As we will show, LPC dramatically increases the separability of different classes and substantially improves robustness to perturbations. 

Another noteworthy effect we observe in our experiments is \textit{binary encoding}, whereby each node in the penultimate layer effectively takes one of two values located symmetrically about the origin. This can be interpreted as class points occupying some vertices of a hypercube defined by the penultimate layer’s dimensions.

\subsection{Related Works}
Our method can be viewed as a form of regularization, but unlike typical approaches such as $L_0$ \cite{louizos2018learning} or $L_1$ \cite{lemhadri2021lassonet} penalties on weights (or dropout \cite{JMLR:v15:srivastava14a}), it compresses the latent representations without changing the network architecture. 
We remark that we propose to add an $L_2$ loss on the \textit{latent representations} and not on the weights of the network, that is already a common regularization practice also known as weight decay.
To our knowledge, the only work that employs an $L_2$ loss
on the penultimate-layer latent representations is the unconstrained feature model developed in \cite{NEURIPS2021_geometricAnalysis},which was used to analyze the global optimization landscape
of the cross-entropy loss function. This theoretical work
included penalties on both the weights and features of the
layer-peeled model, where the $L_2$ loss on the representation
during training was set to a low value. In fact such a low
loss value did not alter the network performance, and it was
used with the aim to justify mathematically the derivation
of a global optimizer, and it does not address the network
enhancements that arise only under a significant penalty on
the $L_2$ loss.
Differently, what we document in this work
is the emergence of LPC, that is a not trivial effect of the
contrast between a strong $L_2$ loss and the standard cross-entropy loss function. This novel concept is elaborated below,
and we show its practical utility especially when applied
onto a low dimensional penultimate linear layer.

Adding loss terms to intermediate layers has also appeared in the context of deep supervision \cite{lee2014deeplysupervised, li2022comprehensive}, where intermediate outputs are trained to match target labels. However, our approach differs by seeking to compress the volume of latent representations rather than providing additional supervision signals.

Various methods have been devised to enlarge inter-class margins, such as contrastive learning \cite{wu2018unsupervisedfeaturelearningnonparametric, hénaff2020dataefficientimagerecognitioncontrastive, oord2019representationlearningcontrastivepredictive, tian2020contrastivemultiviewcoding, hjelm2019learningdeeprepresentationsmutual, chen2020simpleframeworkcontrastivelearning, he2020momentumcontrastunsupervisedvisual} and supervised contrastive learning (SupCon) \cite{khosla2021supervisedcontrastivelearning}, which pull together positive samples while pushing apart negative ones. Other techniques alter the loss function to reduce intra-class variance \cite{centerl} or impose angular constraints \cite{liu2017largemarginsoftmaxlossconvolutional, liu2018spherefacedeephypersphereembedding}, e.g., CosFace \cite{wang2018cosfacelargemargincosine} and ArcFace \cite{Deng_2022}. These last two methods, ArcFace and CosFcae, can be compared to our method in their simplicity, as they each introduce a single penalty term to the loss function to increase margins.

Our approach also makes the network Lipschitz continuous. Prior works on Lipschitz neural networks often rely on architectural constraints such as spectral norm regularization \cite{miyato2018virtualadversarialtrainingregularization, cisse2017parsevalnetworksimprovingrobustness, tsuzuku2018lipschitzmargintrainingscalablecertification}, orthogonal weight matrices \cite{trockman2021orthogonalizingconvolutionallayerscayley, singla2021skeworthogonalconvolutions, singla2022improveddeterministicl2robustness}, or norm-bound weights \cite{anil2019sortinglipschitzfunctionapproximation, gouk2020regularisationneuralnetworksenforcing}, which can reduce model expressiveness or be computationally expensive. By contrast, our method imposes no specialized architectural constraints.

Relatedly, \cite{haas2023linking}  projects latent representations onto a hypersphere surface accelerating convergence to NC. However, they do not induce a full collapse of all representations in each class to a single point, nor do they report notable performance gains beyond improved OOD detection.

Finally, the emergence of LPC creates an information bottleneck (IB) in the latent space, connecting this phenomenon to IB optimization in DNNs \cite{tishby2015deep, alemi2019deepvariationalinformationbottleneck, Kolchinsky_2019, chalk2016relevantsparsecodesvariational, achille2017informationdropoutlearningoptimal, belghazi2021minemutualinformationneural}. Unlike many IB methods, which rely on variational approximations or noise injection, LPC implements a deterministic form of compression through a strong $L_2$ penalty on the features themselves, effectively shrinking their distribution and lowering their entropy.

\section{Method}
\begin{figure}[t]
    \centering
    \includegraphics[width=1.1\columnwidth]{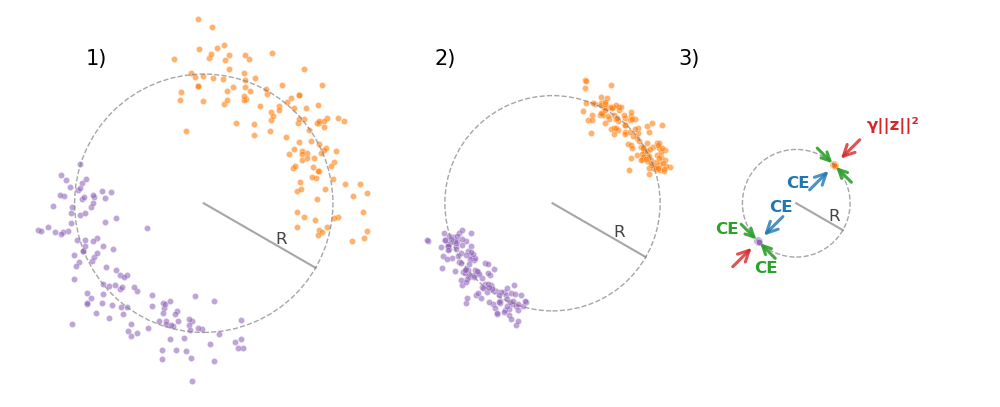}
    \caption{Visualization of the LPC phenomenon with increasing regularization coefficient $\gamma$. (1) We assume class separation in the TPT, meaning that the latent representations of different classes are linearly separable. At low values of $\gamma$, representations are still pushed towards the origin but remain spread around a distance $R$ from it. (2) As $\gamma$ increases, both the radius $R$ and the spread begin to decrease. (3) At high $\gamma$, points collapse to distinct locations while maintaining class separation. The arrows illustrate the competing forces: the blue arrows represent the cross-entropy term driving class separation from the origin, tending to increase the norm of latent representations; the red arrows indicate the compression force ($\gamma\|\mathbf{z}\|^2$) pulling points inward; and the green arrows show the cross-entropy effects pushing different classes apart along the shell, thus promoting convergence within each class. This graphical illustration was created using synthetic data to demonstrate the development of the LPC phenomenon, which is described in detail in App.~\ref{App:latent_point_collapse}.}
    \label{fig:latent_collapse}
\end{figure}

Given a labeled dataset $ \left\{\boldsymbol{x_i},\overline{y_i}\right\},\thinspace i=1\dots N$, where $N$ is the number of data points, we address the problem of predicting the labels using a classifier. 
We utilize a deep neural network that produces a nonlinear mapping of the input $\boldsymbol{f}(\boldsymbol{x})$, aiming to approximate the distribution represented by the data.
Deep neural networks comprise multiple layers stacked together. Each layer produces an internal latent representation. The final output of the network can be expressed as a composition of the functions represented by these layers
$
\boldsymbol{f}(\boldsymbol{x}) = \boldsymbol{f}^{(M)}\circ
\boldsymbol{f}^{(M-1)} \circ \dots \boldsymbol{f}^{(1)}(\boldsymbol{x})
$
where $M$ denotes the number of layers in the network.

For an input vector $\boldsymbol{x}$, the process of generating the neural network's output can be divided, for the sake of this exposition, into two main steps. Firstly, the nonlinear components of the neural network transform the input into a latent representation, denoted as $\boldsymbol{h}(\boldsymbol{x})$. This representation is the output of the last hidden layer before classification.
The final output is then obtained by applying a linear classifier to this latent representation:
$
\boldsymbol{f}(\boldsymbol{x})=\boldsymbol{W}\boldsymbol{h}(\boldsymbol{x}) + \boldsymbol{b}
$
where $\boldsymbol{W}$ and $\boldsymbol{b}$ are the weight matrix and bias vector of the linear classifier, respectively.
The predicted label $y$ is computed by applying a softmax function to the network's output. The softmax function transforms the linear classifier's output into a probability distribution over classes, indicating the likelihood that the input vector $\boldsymbol{x}$ belongs to each class.
The neural network is trained by minimizing the cross-entropy loss function 
$
\mathcal{L}_\text{CE}(\boldsymbol{f}(\boldsymbol{x}), \overline{y})=-\log{\frac{e^{\boldsymbol{f}_{\overline{y}}(\boldsymbol{x})}}{\sum_i e^{\boldsymbol{f}_{i}(\boldsymbol{x})}}}
$
which quantifies the discrepancy between the network's predicted probabilities and the true labels.

We propose to introduce an additional linear layer prior to the classifier, defined as
$
\boldsymbol{z}=\boldsymbol{W}_{\textrm{L2}}\thinspace\boldsymbol{h}(\boldsymbol{x}) + \boldsymbol{b}_{\textrm{L2}}.     
$
This layer serves as the penultimate step in the network architecture, with classification being subsequently determined through another linear operation, 
$
\boldsymbol{f}(\boldsymbol{x})=\boldsymbol{W}\boldsymbol{z} + \boldsymbol{b}.
$
In addition to the cross-entropy loss applied to the network's output, we incorporate an $L_2$ loss function into the penultimate layer defined as:
$
\mathcal{L}_{\textrm{2}}(\mathbf{z})=\|\boldsymbol{z}\|^2,    
$
where $\|\cdot\|$ is the Eucledian norm. 
The resulting loss function is thus composed of two terms
$\mathcal{L} = \mathcal{L}_\textrm{CE} + \gamma\thinspace\mathcal{L}_{\textrm{L2}},$
where $\gamma$ is a positive scalar. 
The latent point collapse emerges as a result of balancing two conflicting tendencies coming from the two components of the loss function

\begin{equation}
\label{eq:loss_full}
\mathcal{L} = -\log{\frac{e^{(\boldsymbol{W}\boldsymbol{z} + \boldsymbol{b})_{\overline{y}}}}{\sum_i e^{(\boldsymbol{W}\boldsymbol{z} + \boldsymbol{b})_i}}} +\gamma \|\boldsymbol{z}\|^2.
\end{equation}

The squared loss function encourages latent representations to be closer to zero, whereas the cross-entropy loss necessitates that the latent representations of different classes be linearly separable in the penultimate layer. 
This configuration generates two opposing tensions: on the one hand, the squared loss drives representations towards each other, potentially making them numerically inseparable or even overlapping; on the other hand, the cross-entropy loss enhances separability and increases the relative distance between latent representations of different classes. 

In App \ref{App:latent_point_collapse} we provide principled arguments that describe how the interplay between these opposing tensions induces LPC that is the collapse of same-class latent representations into a single point, where all collapse points are located at the same distance from the origin. Additionally, in Fig. \ref{fig:latent_collapse}, we provide a graphical illustration demonstrating how an increasing value of the compressing term induces LPC.

Single point collapse of the penultimate-layer representations ensures Lipschitz continuity of the network defined as 
\begin{equation}
\|\boldsymbol{f}(\boldsymbol{x}_1)-\boldsymbol{f}(\boldsymbol{x}_2)\|\leq L\|\boldsymbol{x}_1-\boldsymbol{x}_2\|.     
\end{equation}
Considering that the final output representations are obtained as a linear combination of the latent representations $\boldsymbol{z}$ which are located in the sorrounding of the origin, we can conclude that LPC renders networks Lipschitz.
More specifically, in \cite{zhang2022rethinkinglipschitzneuralnetworks, li2019preventinggradientattenuationlipschitz}, it is shown that the minimum amount of perturbation required to induce misclassification in a Lipschitz network is proportional to the Lipschitz constant $L$, and inversely proportional to the margin of $\boldsymbol{f}(\boldsymbol{x})$, i.e., the distance from the decision boundary. In the limit of a latent point collapse, $L$ and the margin of $\boldsymbol{f}(\boldsymbol{x})$ are both proportional to distance of the collapse points from the origin. Thus, it is possible to set a minimum amount of perturbation that is necessary to trigger a misclassification.
The same reasoning cannot be done in case of networks whose penultimate latent representations are not bounded close to the origin.

\subsection{Binary encoding}

\begin{figure}[t]
\vskip 0.1in
\begin{center}
\centerline{\includegraphics[width=3.3in]{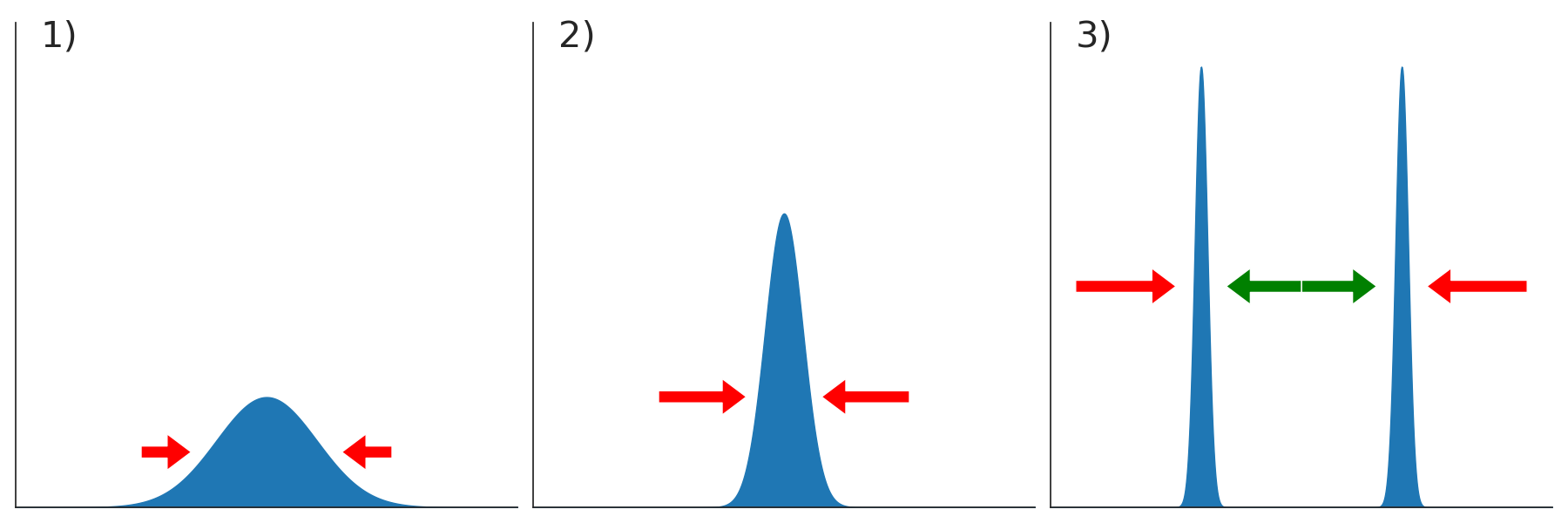}}
\caption{
Graphical illustration of the dynamics leading to the emergence of a latent binary encoding. 
The three images provide a qualitative representation of the training process, where the scalar $\gamma$ is progressively increased. 
The plots in the images represent histograms of the latent representations at a specific node of the linear penultimate layer. 
In the first image, the relatively low value of $\gamma$ constrains all values close to the origin, but the volume remains large enough for the network to differentiate between different classes. 
As $\gamma$ increases, all latent values are drawn closer to the origin, as depicted in the second image, making it increasingly difficult for the network to discriminate between elements of different classes. 
Consequently, the network is forced to find a more stable solution through numerical optimization, placing all elements belonging to the same class in the neighborhood of one of two points. 
In the two distributions shown in the third image, each of the two peaks contains elements from different classes, but all elements of a specific class are confined to a single peak. 
In other words, while a peak may contain multiple classes, all elements of the same class are restricted to the same peak. 
Through numerical optimization, these two peaks eventually converge to single points, positioned opposite to each other with respect to the origin, as illustrated in the third image—an outcome facilitated by the linear layer. 
The red (green) arrow represents the net effect of the binary encoding (cross-entropy) loss. 
In relation to Fig.~\ref{fig:latent_collapse}, this figure empirically demonstrates that the points of collapse are located on some vertices of a hypercube defined by the dimensions of the penultimate layer, which explains the binary structure observed in this figure.
}

\label{Fig:illustration} 
\end{center}
\vskip -0.1in
\end{figure}

The points of collapse are located on a hypersphere, and our experiments demonstrate that they align with the vertices of a hypercube inscribed within this hypersphere. 
More precisely, we find that at each node of the penultimate layer, latent representations can approximately assume one of two values, thereby forming a binary encoding. 
However, we do not provide an explanation for why these collapse points specifically correspond to the vertices of a hypercube. 
One possible argument is that as latent representations approach the origin, it becomes increasingly difficult for the network to accurately position all representations within such a confined space without mistakenly placing a representation at a vertex associated with a different class. 
A practical solution to this challenge, potentially discovered by the optimizer, is to maximize the relative distance between different collapse points  {\em in each dimension} of the latent space. 
This can be achieved by arranging opposing groups of collapse points symmetrically around the origin, as illustrated in Fig. \ref{Fig:illustration}.
We note that such a symmetric arrangement can be realized using a linear layer, which allows for the symmetric displacement of latent representations relative to the origin. 

\subsection{Information bottleneck}
Our loss function induces the collapse of all same-class latent representations into a single point, which can also be posed as a method to create an IB in the penultimate layer.
The optimization of the IB Lagrangian aims to maximize the following objective:

\begin{equation}
\label{eq:loss_ib}
\mathcal{L}_{IB} = I(\mathbf{z}; \overline{y}) - \beta I(\mathbf{z}; \mathbf{x}),
\end{equation}

where \( I(\mathbf{z}; \overline{y}) \) denotes the mutual information between the latent representation \( \mathbf{z} \) and the labels \( \overline{y} \) and \( I(\mathbf{z}; \mathbf{x}) \) represents the mutual information between \( \mathbf{z} \) and the input data \( \mathbf{x} \). The parameter \( \beta \) controls the trade-off between compression and predictive accuracy.
In App. \ref{App:IB}, we demonstrate that minimizing this quantity in deterministic DNN classification is equivalent to minimizing the following quantity:

\begin{equation}
\label{eq:loss_ib_H_CE}
\mathcal{L}_{IB} =\mathcal{L}_\text{CE}(\boldsymbol{f}(\boldsymbol{x}), \overline{y}) - \beta  H(\mathbf{z}),
\end{equation}

where $H(\mathbf{z})$ is the entropy associated with the latent distribution $\mathbf{z}$ and $\mathcal{L}_\text{CE}(\boldsymbol{f}(\boldsymbol{x})$ is the cross-entropy loss function. 
During training, the cross-entropy loss function is directly minimized, while the entropy $H(\mathbf{z})$ is indirectly minimized by the collapse of all same-class latent representations into a single point. 

In order to understand how LPC effectively minimizes the entropy associated with the probability distributions that generate the latent representations $\mathbf{z}$, we approximate the differential entropy with a discrete Shannon entropy and take the limit for an infinitesimally small quantization.

\begin{equation}
    H_{\Delta} = -\sum_i p_i \log p_i
\end{equation}

As a result of the collapse of all same-class latent representations into a single point, all elements of a specific class are confined to a unique bin, even in the limit of a very small bin size. 
In case of $K$ classes where each class contains the same number of elements, the entropy reduces to:

\begin{equation}
    H_{\Delta} = - \log\frac{1}{K}.
\end{equation}

This represents the minimum possible value of entropy that still permits discrimination among classes.
If the latent representations do not collapse into a single point, the distribution will be spread across multiple bins, resulting in a higher entropy.

\section{Experiments}
The aim of these experiments is to empirically demonstrate that our method promotes latent point collapse in the penultimate layer of DNN classifiers and to show how incorporating a low-dimensional penultimate linear layer can improve network performance. We conduct an ablation study to assess the impact of each property characterizing the penultimate layer as introduced in our method.

All architectures share a common backbone that generates the latent representation $\mathbf{h}(\mathbf{x})$, but differ in how they carry out the final classification. We begin with a network that incorporates a linear penultimate layer with an $L_2$ loss applied to its latent representation, experimenting with three different dimensionalities for this layer: low, intermediate, and high. To evaluate the effect of the $L_2$ loss, we include two variations with an intermediate-dimensional layer (one linear and one non-linear), both trained without the $L_2$ loss.

To assess the impact of adding a low-dimensional linear penultimate layer, we include an architecture that applies an $L_2$ loss directly to the backbone—without a separate penultimate layer—before classification. Additionally, we include architectures that implement the SupCon, ArcFace, and CosFace loss functions. Finally, we test a variant that combines the $L_2$ loss on an intermediate penultimate linear layer with SupCon on the backbone’s latent representations; we note that these two losses operate on different layers.

We refer to the architecture with an $L_2$ loss applied to a linear penultimate layer of intermediate dimensionality as \textsc{LPC}, the lower-dimensional model as \textsc{LPC-Narrow}, and the higher-dimensional model as \textsc{LPC-Wide}. The architecture featuring an $L_2$ loss on the penultimate layer and SupCon on the backbone is called \textsc{LPC-SCL}. The \textsc{NoPenLPC} model applies an $L_2$ loss directly to the backbone and, unlike the other models, does not include an intermediate layer. The linear penultimate (\textsc{LinPen}) and non-linear penultimate (\textsc{NonlinPen}) architectures both have a penultimate layer (linear or non-linear, respectively) with the same dimensionality as \textsc{LPC}, but are trained solely using cross-entropy loss. The no-penultimate (\textsc{NoPen}) model serves as the baseline architecture, performing linear classification directly on $\mathbf{h}(\mathbf{x})$ and training only with cross-entropy loss. The \textsc{SCL}, \textsc{ArcFace}, and \textsc{CosFace} architectures implement SCL, ArcFace, and CosFace losses, respectively, on this baseline.

All experiments were conducted on the CIFAR10 and CIFAR100 datasets \cite{cifar}, using ResNet architectures \cite{resnet}. Code to reproduce our results is available online in the linked repository.\footnote{\href{https://github.com/luigisbailo/lpc}{https://github.com/luigisbailo/lpc}} All experimental details, along with a summary of the different architectures used, are provided in Appendix~\ref{App:training_details}. We note that $L_2$ weight decay was used consistently in training all architectures.
{
\setlength{\tabcolsep}{10pt}
\begin{table*}[t]
\caption{All values in the table represent means and standard deviations from different experiments. 
\textit{Left column}: class separation ratio ($\mathcal{R}$); 
\textit{Center-Left column}: within-class covariance ($\Sigma_W$); 
\textit{Center-Right column}: coefficient of variation (std./mean) of the latent-norm distribution; 
\textit{Right column}: estimated entropy ($H$) using the Kozachenko-Leonenko method (k=20), 
divided by the penultimate-layer dimension.}
\label{table:point_collapse}
\vskip 0.2in
\begin{center}
\begin{small}
\begin{sc}
\begin{tabular}{lcccc}
\toprule
\multicolumn{5}{c}{\textbf{Dataset: CIFAR10}} \\
\cmidrule(lr){1-5}
\textbf{Model}
 & $\boldsymbol{\mathcal{R}}$
 & $\boldsymbol{\Sigma_W}$
 & \textbf{Coeff. of Var.}
 & $\boldsymbol{H}$ \\
\midrule
LPC
 & $\phantom{00}298.61 \,\pm\, 58.09$
 & $1.1064\times 10^{-13} \pm 2.7650\times 10^{-14}$
 & $0.012 \pm 0.004$
 & $-4.123 \pm 0.200$ \\

LPC-Wide
 & $\phantom{00}238.67 \,\pm\, 24.79$
 & $5.7068\times 10^{-14} \pm 1.4499\times 10^{-14}$
 & $0.016 \pm 0.004$
 & $-3.946 \pm 0.113$ \\

LPC-Narrow
 & $\phantom{00}359.89 \,\pm\, 77.09$
 & $5.8526\times 10^{-13} \pm 2.3445\times 10^{-13}$
 & $0.015 \pm 0.012$
 & $-4.065 \pm 0.262$ \\

LPC-NoPen
 & $\phantom{000}28.06 \,\pm\, \phantom{0}6.73$
 & $6.7506\times 10^{-12} \pm 7.9004\times 10^{-12}$
 & $0.016 \pm 0.003$
 & $-1.337 \pm 0.252$ \\

LPC-SCL
 & $\phantom{00}184.57 \,\pm\, \phantom{0}5.21$
 & $2.3933\times 10^{-13} \pm 6.9241\times 10^{-14}$
 & $0.015 \pm 0.002$
 & $-3.678 \pm 0.035$ \\

LinPen
 & $\phantom{000}\,3.12 \,\pm\, \phantom{0}0.03$
 & $7.9841\times 10^{-2} \pm 3.7791\times 10^{-2}$
 & $0.359 \pm 0.054$
 & $\phantom{-}0.213 \pm 0.044$ \\

NonlinPen
 & $\phantom{000}\,3.25 \,\pm\, \phantom{0}0.11$
 & $1.8192\times 10^{-1} \pm 2.6239\times 10^{-2}$
 & $0.268 \pm 0.039$
 & $\phantom{-}0.253 \pm 0.026$ \\

SCL
 & $\phantom{00}11.20 \,\pm\, \phantom{0}0.37$
 & $8.9445\times 10^{-3} \pm 3.0380\times 10^{-3}$
 & $0.228 \pm 0.014$
 & $-0.170 \pm 0.413$ \\

ArcFace
 & $\phantom{000}\,5.51 \,\pm\, \phantom{0}0.18$
 & $8.5891\times 10^{-2} \pm 1.8290\times 10^{-2}$
 & $0.554 \pm 0.044$
 & $-0.268 \pm 0.057$ \\

CosFace
 & $\phantom{000}\,3.66 \,\pm\, \phantom{0}0.13$
 & $7.1308\times 10^{-2} \pm 5.8374\times 10^{-3}$
 & $0.475 \pm 0.023$
 & $\phantom{-}0.115 \pm 0.082$ \\

NoPen
 & $\phantom{000}\,1.91 \,\pm\, \phantom{0}0.05$
 & $3.4388\times 10^{-2} \pm 5.6878\times 10^{-3}$
 & $0.363 \pm 0.055$
 & $\phantom{-}0.722 \pm 0.029$ \\
\addlinespace
\midrule
\multicolumn{5}{c}{\textbf{Dataset: CIFAR100}} \\
\cmidrule(lr){1-5}
\textbf{Model}
 & $\boldsymbol{\mathcal{R}}$
 & $\boldsymbol{\Sigma_W}$
 & \textbf{Coeff. of Var.}
 & $\boldsymbol{H}$ \\
\midrule
LPC
 & $\phantom{00}194.95 \,\pm\, \phantom{0}4.13$
 & $1.3084\times 10^{-12} \pm 5.5604\times 10^{-13}$
 & $0.006 \pm 0.000$
 & $-3.471 \pm 0.021$ \\

LPC-Wide
 & $\phantom{00}140.24 \,\pm\, \phantom{0}1.11$
 & $6.7149\times 10^{-13} \pm 2.4117\times 10^{-13}$
 & $0.009 \pm 0.000$
 & $-3.092 \pm 0.007$ \\

LPC-Narrow
 & $\phantom{00}213.73 \,\pm\, 22.10$
 & $2.8936\times 10^{-12} \pm 1.2199\times 10^{-12}$
 & $0.010 \pm 0.001$
 & $-3.587 \pm 0.099$ \\

LPC-NoPen
 & $\phantom{000}\,6.29 \,\pm\,  \phantom{0}4.73$
 & $7.6487\times 10^{-6}  \pm 8.1990\times 10^{-6}$
 & $7.989 \pm 4.091$
 & $\phantom{-}1.873 \pm 1.997$ \\

LPC-SCL
 & $\phantom{00}109.96 \,\pm\, 13.42$
 & $1.9884\times 10^{-12} \pm 1.1550\times 10^{-12}$
 & $0.011 \pm 0.000$
 & $-2.896 \pm 0.111$ \\

LinPen
 & $\phantom{000}\,1.49 \,\pm\,  \phantom{0}0.00$
 & $1.3637\times 10^{-1} \pm 2.6163\times 10^{-2}$
 & $0.252 \pm 0.003$
 & $\phantom{-}1.087 \pm 0.004$ \\

NonlinPen
 & $\phantom{000}\,1.48 \,\pm\,  \phantom{0}0.03$
 & $1.1608 \pm 0.3729$
 & $0.250 \pm 0.011$
 & $\phantom{-}1.075 \pm 0.026$ \\

SCL
 & $\phantom{000}\,8.48 \,\pm\,  \phantom{0}0.48$
 & $1.1722\times 10^{-3} \pm 1.8947\times 10^{-3}$
 & $0.200 \pm 0.012$
 & $\phantom{-}5.424 \pm 0.022$ \\

ArcFace
 & $\phantom{000}\,4.94 \,\pm\,  \phantom{0}0.09$
 & $4.6705\times 10^{-2} \pm 3.5804\times 10^{-3}$
 & $0.354 \pm 0.012$
 & $\phantom{-}5.378 \pm 0.022$ \\

CosFace
 & $\phantom{000}\,3.66 \,\pm\,  \phantom{0}0.14$
 & $5.0610\times 10^{-2} \pm 1.5943\times 10^{-2}$
 & $0.425 \pm 0.024$
 & $\phantom{-}5.227 \pm 0.009$ \\

NoPen
 & $\phantom{000}\,1.21 \,\pm\,  \phantom{0}0.00$
 & $3.4299\times 10^{-2} \pm 3.8287\times 10^{-3}$
 & $0.293 \pm 0.017$
 & $\phantom{-}5.547 \pm 0.018$ \\
\bottomrule
\end{tabular}
\end{sc}
\end{small}
\end{center}
\vskip -0.15in
\end{table*}
}

\subsection{Latent Point Collapse}

To investigate whether a latent point collapse occurs in the penultimate layer, we examine the within-class covariance $\Sigma_{W}$, defined as
\begin{equation}
\label{eq:sigma_w}
\Sigma_{W}=\frac{1}{NP}\sum_{i=0}^{N-1}\sum_{p=0}^{P-1}\left(\boldsymbol{z}^{(i,p)} - \boldsymbol{\mu}^{(p)}\right)\left(\boldsymbol{z}^{(i,p)} - \boldsymbol{\mu}^{(p)}\right)^\top,
\end{equation}
where $\boldsymbol{z}^{(i,p)}$ is the $i$-th latent representation with label $p$, and $\boldsymbol{\mu}^{(p)}$ is the mean of all latent representations with label $p$.

To assess separability in latent space, we define a \emph{class separation ratio}, $\mathcal{R}$, as the ratio of the distance between a latent point $\boldsymbol{z}^{(i,p)}$ and the nearest \emph{other-class} centroid, $\boldsymbol{\mu}^{(q)}$ ($q \neq p$), to its distance from its \emph{own-class} centroid, $\boldsymbol{\mu}^{(p)}$:

\begin{equation}
\label{eq:margin}
\mathcal{R}^{(i,p)} = \frac{\min_{q \neq p} \|\boldsymbol{z}^{(i,p)} - \boldsymbol{\mu}^{(q)}\|}{\|\boldsymbol{z}^{(i,p)} - \boldsymbol{\mu}^{(p)}\|}.
\end{equation}

A higher value of this ratio indicates better class separation. We report the average across all samples as a metric.

We also test whether all collapse points lie at the same distance from the origin, as discussed in Sec.~\ref{App:latent_point_collapse}. Additionally, we estimate the entropy of the latent distribution in the penultimate layer using the Kozachenko--Leonenko $k$-nearest neighbors method.

Table~\ref{table:point_collapse} presents the values of the above quantities at the final epoch of training for the various architectures considered. Notably, $\Sigma_W$ converges to zero only for architectures that implement an $L_2$ loss on the penultimate layer, indicating that all same-class latent representations collapse to a single point.

We also observe that these architectures achieve a class separation ratio more than two orders of magnitude larger than the baseline method. In contrast, other methods we compare against---SCL, ArcFace, and CosFace---improve the class separation ratio by less than an order of magnitude relative to the baseline.

Furthermore, we analyze the coefficient of variation of the distance from the origin to each collapse point and observe that these points consistently remain at an approximately constant distance from the origin. Lastly, the entropy of the distribution sharply decreases due to the point collapse, which confines all points within a small volume.

In our experiments, we show that the collapse points lie at the vertices of a hypercube, as discussed in Appendix~\ref{App:binarity_hypothesis}. Additionally, in Appendix~\ref{App:neural_collapse}, we compare this single-point collapse with the NC phenomenon. In particular, we note that our training was largely performed in the TPT regime, ensuring that the improvements we observe occur in addition to those typically associated with NC.

\subsection{Robustness and generalization}
\begin{table}[t]
\caption{All values in the table represent the means and standard deviations obtained from different experiments. 
\textit{Left column}:  Robustness of the network computed as the norm of the minimal amount of perturbation, divided by the norm of the input, to cause a prediction change. The algorithm describing the method to produce the perturbation is in \cite{deepfool}. We present in this table the average results across the different experiments, where in each experiment the algorithm method was tested on 1000 inputs sampled from the testing set. 
\textit{Right column}: lassification accuracy on the testing set at the last epoch.}
\label{table:results}
\vskip 0.15in
\begin{center}
\begin{small}
\begin{sc}
\begin{tabular}{lcc}
\toprule
\multicolumn{3}{c}{\textbf{Dataset: CIFAR10}} \\
\cmidrule(lr){1-3}
\textbf{Model} & \textbf{Robustness} & \textbf{Accuracy} \\
\midrule
LPC         & $0.586 \pm 0.175$      & $\mathbf{94.98} \pm 0.12$ \\
LPC-Wide    & $0.479 \pm 0.063$      & $94.86 \pm 0.12$ \\
LPC-Narrow  & $\mathbf{0.731} \pm 0.119$ & $94.77 \pm 0.09$ \\
LPC-NoPen   & $0.240 \pm 0.044$      & $94.69 \pm 0.12$ \\
LPC-SCL     & $0.347 \pm 0.017$      & $94.75 \pm 0.09$ \\
LinPen      & $0.013 \pm 0.000$      & $94.54 \pm 0.07$ \\
NonlinPen   & $0.016 \pm 0.002$      & $94.46 \pm 0.07$ \\
SCL         & $0.028 \pm 0.002$      & $94.62 \pm 0.15$ \\
ArcFace     & $0.019 \pm 0.001$      & $94.64 \pm 0.18$ \\
CosFace     & $0.018 \pm 0.001$      & $94.56 \pm 0.10$ \\
NoPen       & $0.013 \pm 0.000$      & $94.50 \pm 0.14$ \\
\addlinespace
\midrule
\multicolumn{3}{c}{\textbf{Dataset: CIFAR100}} \\
\cmidrule(lr){1-3}
\textbf{Model} & \textbf{Robustness} & \textbf{Accuracy} \\
\midrule
LPC         & $0.183 \pm 0.012$      & $77.96 \pm 0.26$ \\
LPC-Wide    & $0.167 \pm 0.007$      & $77.78 \pm 0.19$ \\
LPC-Narrow  & $\mathbf{0.227} \pm 0.012$ & $77.42 \pm 0.20$ \\
LPC-NoPen   & $0.054 \pm 0.004$      & $76.88 \pm 0.19$ \\
LPC-SCL     & $0.129 \pm 0.017$      & $\mathbf{78.15} \pm 0.05$ \\
LinPen      & $0.007 \pm 0.000$      & $76.79 \pm 0.08$ \\
NonlinPen   & $0.007 \pm 0.000$      & $76.68 \pm 0.12$ \\
SCL         & $0.012 \pm 0.001$      & $77.96 \pm 0.14$ \\
ArcFace     & $0.013 \pm 0.000$      & $77.29 \pm 0.35$ \\
CosFace     & $0.010 \pm 0.000$      & $76.89 \pm 0.26$ \\
NoPen       & $0.007 \pm 0.000$      & $77.22 \pm 0.41$ \\
\bottomrule
\end{tabular}
\end{sc}
\end{small}
\end{center}
\vskip -0.1in
\end{table}

\begin{figure}[ht]
\vskip 0.2in
\begin{center}
\centerline{\includegraphics[width=\columnwidth]{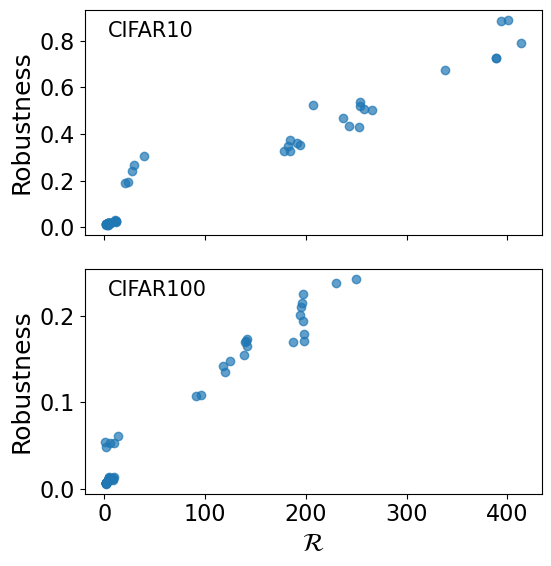}}
\caption{Each point represents a different training instance, and points are taken from all evaluated architectures. As shown in the plot, there is a strong correlation between the class separation ratio $\mathcal{R}$ and the network’s robustness to input perturbations. The Pearson correlation coefficients are 0.97 for CIFAR10 and 0.985 for CIFAR100, respectively.
}
\label{Fig:margins}
\end{center}
\vskip -0.2in
\end{figure}

Results in Table \ref{table:results} show the magnitude of the minimal perturbation on the input data required to change the classification label. To obtain this value, we employed the DeepFool algorithm \cite{deepfool}. In our experiments, we observe a dramatic improvement in the network's robustness when an $L_2$ loss is applied to a penultimate layer---over two orders of magnitude of improvement, specifically when the penultimate layer is linear. The other regularization techniques (SCL, ArcFace, CosFace) also improve robustness relative to the baseline, but their gains are not as pronounced. We attribute the increased robustness to increased separability, as illustrated in Fig. \ref{Fig:margins}, where larger class separation directly correlate with improved robustness.

We also note that LPC has an additional regularizing effect on the network and thus improves generalization. Its generalization performance is comparable to, or even surpasses, state-of-the-art regularization methods such as SCL. Interestingly, combining LPC with SCL yields the best generalization results on CIFAR100, demonstrating how LPC can be combined with other known regularization techniques.

Our experiments further reveal that adding a linear penultimate layer is practically beneficial. We find that an intermediate-dimensional penultimate layer tends to provide better generalization, whereas a lower-dimensional penultimate layer yields better robustness. In any case, applying $L_2$ regularization on a linear penultimate layer---rather than directly on the backbone---consistently results in large improvements in both robustness and generalization.

\section{Discussion}

The LPC phenomenon greatly increases the separability of latent representations, leading to a dramatic improvement in robustness. This effect arises from applying a strong $L_2$ loss on the penultimate layer, and we show that it is practically useful to apply such a loss to a linear penultimate layer with a dimensional bottleneck. We also demonstrate how LPC can be combined with other regularization techniques.

These improvements in robustness and generalization also occur in the TPT and are associated with the NC phenomenon. However, our experimental design ensures that a significant portion of training is conducted in the TPT, implying that our observed benefits are in addition to those naturally arising from NC.

The LPC phenomenon we describe results from the interplay between cross-entropy loss and $L_2$ regularization on the penultimate layer. The different asymptotic behaviors of these two functions give rise to an equilibrium point, leading to point collapse. Since NC has been shown to develop under a variety of loss functions beyond cross-entropy \cite{han2022neural,zhou2022lossescreatedequalneural}, it would be interesting to explore similar interactions with those alternative losses.

\subsection{Conclusion}
In this paper, we introduced a straightforward method for inducing the collapse of latent representations onto the vertices of a hypercube. We demonstrated that this approach significantly enhances network robustness while also yielding a small but statistically significant improvement in generalization. Additionally, we showed the practical benefits of inserting a low-dimensional linear layer between the backbone and the final linear classifier as the penultimate layer. 
The simplicity of this method is striking, making it particularly appealing for practical applications.

\section{Acknowledgements}
This research is supported by the project "Memristors Materials by Design, MeMabyDe", by the Leibniz Association.
The authors gratefully acknowledge the scientific support and HPC resources provided by the Erlangen National High Performance Computing Center (NHR@FAU) of the Friedrich-Alexander-Universität Erlangen-Nürnberg (FAU). The hardware is funded by the German Research Foundation (DFG).

\newpage

\newpage
\bibliography{main}

\newpage
\appendix

\section{Derivation of Latent Point Collapse}
\label{App:latent_point_collapse}

In this appendix, we provide a comprehensive derivation showing that imposing a strong $L_2$ loss on the penultimate-layer latent representations causes the latent representations of each class to collapse to a single point.

We consider a deep neural network classifier, where the penultimate-layer representations are denoted by \(\boldsymbol{z}\in\mathbb{R}^d\).  A linear classifier with weight matrix \(\boldsymbol{W}\in\mathbb{R}^{K\times d}\) (for \(K\) classes) and bias \(\boldsymbol{b}\in\mathbb{R}^K\) produces logits
\[
(\boldsymbol{W}\boldsymbol{z} + \boldsymbol{b})_i
\quad
\text{for}
\quad
i=1,\dots,K.
\]
We study the loss function
\begin{equation}
    \label{eq:full_loss}
    \mathcal{L}(\boldsymbol{z})
    \;=\;
    \mathcal{L}_{\text{CE}}(\boldsymbol{z})
    \;+\;
    \gamma\,\|\boldsymbol{z}\|^2,
\end{equation}
where \(\mathcal{L}_{\text{CE}}\) is the cross-entropy loss and \(\gamma>0\) is the regularization coefficient on the latent representation.

For a sample \(\boldsymbol{z}\) with ground-truth label \(\overline{y}\), the cross-entropy loss is
\[
    \mathcal{L}_{\text{CE}}(\boldsymbol{z})
    \;=\;
    -\;\log
        \biggl[
             \frac{e^{(\boldsymbol{W}\boldsymbol{z}+\boldsymbol{b})_{\overline{y}}}}%
             {\sum_{j} e^{(\boldsymbol{W}\boldsymbol{z}+\boldsymbol{b})_{j}}}
        \biggr].
\]
Let
\[
    p_i
    \;=\;
    \frac{e^{(\boldsymbol{W}\boldsymbol{z}+\boldsymbol{b})_{i}}}%
             {\sum_{j} e^{(\boldsymbol{W}\boldsymbol{z}+\boldsymbol{b})_{j}}}
\]
be the softmax probability of class \(i\).  The compression term is
\(\mathcal{L}_{\text{2}}(\boldsymbol{z}) = \gamma\,\|\boldsymbol{z}\|^2\),
whose gradient is
\[
    \nabla \mathcal{L}_{\text{2}}(\boldsymbol{z}) = 2\,\gamma\,\boldsymbol{z}.
\]
Combining the gradients of cross-entropy and compression yields
\begin{equation}
    \label{eq:grad_full}
    \nabla \mathcal{L}(\boldsymbol{z})
    \;=\;
    \underbrace{
    \Bigl(
       -\,\boldsymbol{W}_{\overline{y}}
       \;+\;
       \sum_{i=1}^K p_i\,\boldsymbol{W}_i
    \Bigr)}_{\displaystyle \nabla \mathcal{L}_{\text{CE}}(\boldsymbol{z})}
    \;+\;
    \underbrace{
    2\,\gamma\,\boldsymbol{z}
    }_{\displaystyle \nabla \mathcal{L}_{\text{H}}(\boldsymbol{z})}.
\end{equation}

We focus on the terminal phase of training, where the classes are separated but the model has not reached a stationary point. Formally:

\begin{assumption}[Class Separation]
\label{assump:separation}
For every training sample \(\boldsymbol{z}\) with ground-truth label \(\overline{y}\), 
\[
    \boldsymbol{W}_{\overline{y}}^\top \boldsymbol{z}
    \;>\;
    \boldsymbol{W}_{i}^\top \boldsymbol{z}
    \quad \text{for all } i\neq \overline{y}.
\]
\end{assumption}

\begin{assumption}[Non-vanishing Gradients]
\label{assump:gradients}
There exists \(\varepsilon > 0\) such that
\[
    \|\nabla \mathcal{L}(\boldsymbol{z})\|
    \;>\;
    \varepsilon
    \quad
    \text{throughout this training phase.}
\]
\end{assumption}

Assumption~\ref{assump:separation} states that the correct logit \(\boldsymbol{W}_{\overline{y}}^\top \boldsymbol{z}\) is strictly larger than any incorrect logit, while Assumption~\ref{assump:gradients} ensures training has not converged.

\begin{lemma}[Softmax Approximation in the Terminal Phase]
\label{lemma:softmax}
Under Assumption~\ref{assump:separation}, there exists \(\delta>0\) such that
\[
    \boldsymbol{W}_{\overline{y}}^\top\boldsymbol{z} \;-\; \boldsymbol{W}_i^\top\boldsymbol{z}
    \;\ge\; \delta
    \quad 
    \text{for all } i\neq \overline{y}.
\]
Let \(\alpha = e^{-\delta} < 1\). Then for \(i\neq \overline{y}\),
\[
    \exp\!\bigl((\boldsymbol{W}\boldsymbol{z}+\boldsymbol{b})_i\bigr)
    \;\le\;
    \alpha\,
    \exp\!\bigl((\boldsymbol{W}\boldsymbol{z}+\boldsymbol{b})_{\overline{y}}\bigr),
\]
and therefore
\[
    \sum_{j=1}^K
    \exp\!\bigl((\boldsymbol{W}\boldsymbol{z}+\boldsymbol{b})_j\bigr)
    \;=\;
    \exp\!\bigl((\boldsymbol{W}\boldsymbol{z}+\boldsymbol{b})_{\overline{y}}\bigr)\,
    \bigl(1 + \mathcal{O}(\alpha)\bigr).
\]
Hence \(p_{\overline{y}}\approx 1\) and \(p_i\approx 0\) for \(i\neq \overline{y}\) in the terminal phase.
\end{lemma}
Concretely, because \( p_{\overline{y}}(\mathbf{z}) \approx 1\), there is no large gain in further “pushing up” the logit \(\mathbf{W}_{\overline{y}}^\top \mathbf{z}\). As a result, the partial derivatives of the cross-entropy loss cannot grow unboundedly with respect to \(\|\mathbf{z}\|\). Formally, if
$
   \nabla \mathcal{L}_{\mathrm{CE}}(\mathbf{z})
   \;=\;
   -\,\mathbf{W}_{\overline{y}}
   \;+\;
   \sum_{i=1}^K
   p_{i}(\mathbf{z})\,\mathbf{W}_i,
$
then in the region where \(p_{\overline{y}}(\mathbf{z})\approx 1\), we have
\(\sum_{i=1}^K p_i(\mathbf{z})\,\mathbf{W}_i \approx \mathbf{W}_{\overline{y}}\),
so \(\|\nabla \mathcal{L}_{\mathrm{CE}}(\mathbf{z})\|\) remains essentially bounded. This boundedness ensures that, for large norms of \(\mathbf{z}\), the compression gradient \(2\,\gamma\,\mathbf{z}\) dominates, driving \(\mathbf{z}\) back inward.

From \(\mathcal{L}_{\text{H}}(\boldsymbol{z}) = \gamma\,\|\boldsymbol{z}\|^2\), we have
\[
    \nabla \mathcal{L}_{\text{2}}(\boldsymbol{z}) = 2\,\gamma\,\boldsymbol{z}.
\]
Thus:

\begin{lemma}[Radial Flow]
\label{lemma:flow}
For any \(\boldsymbol{z}\neq \mathbf{0}\),
\[
    \langle 
       \boldsymbol{z},
       \,\nabla \mathcal{L}_{\text{2}}(\boldsymbol{z})
    \rangle
    \;=\;
    2\,\gamma\,\|\boldsymbol{z}\|^2
    \;>\;
    0.
\]
\end{lemma}

When descending this loss, the term \(2\,\gamma\,\boldsymbol{z}\) pulls \(\boldsymbol{z}\) radially inward toward the origin.

\subsection{Detailed Radial Analysis Toward Equilibrium}
\label{sec:terminal_phase_equilibrium}

\paragraph{Behavior for large \(\|\boldsymbol{z}\|\).}  
Since \(\|\nabla \mathcal{L}_{\text{2}}(\boldsymbol{z})\|\;=\;2\,\gamma\,\|\boldsymbol{z}\|\) grows linearly in \(\|\boldsymbol{z}\|\), while \(\|\nabla \mathcal{L}_{\text{CE}}(\boldsymbol{z})\|\) remains effectively bounded in the terminal phase (Lemma~\ref{lemma:softmax}), the inward pull from compression dominates if \(\|\boldsymbol{z}\|\) grows too large.  This stabilizes \(\|\boldsymbol{z}\|\) around some finite radius.

\paragraph{Behavior for small \(\|\boldsymbol{z}\|\).}
Conversely, if \(\|\boldsymbol{z}\|\) is too small, the logits become under-confident, increasing cross-entropy loss.  Its gradient near \(\boldsymbol{z}=\mathbf{0}\) is non-zero, pushing \(\boldsymbol{z}\) outward.  Meanwhile, the compression gradient is small near the origin (since it is proportional to \(\|\boldsymbol{z}\|\)).  

\paragraph{Radial derivative along \(\boldsymbol{W}_{\overline{y}}\).}
To analyze how the cross-entropy loss changes when the latent representation \(\boldsymbol{z}\) moves along the direction of the correct class weight \(\boldsymbol{W}_{\overline{y}}\), we consider the directional derivative:

\[
   \bigl\langle
      \nabla \mathcal{L}_{\text{CE}}(\boldsymbol{z}),
      \;\hat{u}
   \bigr\rangle
\]

where \(\hat{u} = \boldsymbol{W}_{\overline{y}} / \|\boldsymbol{W}_{\overline{y}}\|\) is the unit vector in the direction of \(\boldsymbol{W}_{\overline{y}}\). 

Using the gradient of the cross-entropy loss (Eq.~\eqref{eq:grad_full}) we compute:

\[
   \bigl\langle
      \nabla \mathcal{L}_{\text{CE}}(\boldsymbol{z}),
      \;\hat{u}
   \bigr\rangle
   =
   \sum_{i=1}^K p_i(\boldsymbol{z})
      \,\frac{\boldsymbol{W}_{i}^\top \boldsymbol{W}_{\overline{y}}}%
            {\|\boldsymbol{W}_{\overline{y}}\|}
   \;-\;
   \|\boldsymbol{W}_{\overline{y}}\|.
\]

Since \( \boldsymbol{W}_{i}^\top \boldsymbol{W}_{\overline{y}} < \|\boldsymbol{W}_{\overline{y}}\|^2 \) for all \( i \neq \overline{y} \), it follows that:

\[
   \bigl\langle
      \nabla \mathcal{L}_{\text{CE}}(\boldsymbol{z}),
      \;\hat{u}
   \bigr\rangle < 0.
\]

Thus, increasing \( \|\boldsymbol{z}\| \) in the direction of \( \boldsymbol{W}_{\overline{y}} \) reduces the cross-entropy loss.

Moreover, since \( p_{\overline{y}} \to 1 \) and \( p_i \to 0 \) for \( i \neq \overline{y} \) in the terminal phase (Lemma~\ref{lemma:softmax}), the sum:

\[
\sum_{i=1}^K p_i(\boldsymbol{z}) \,\frac{\boldsymbol{W}_{i}^\top \boldsymbol{W}_{\overline{y}}}{\|\boldsymbol{W}_{\overline{y}}\|}
\]

shrinks in magnitude as \( \|\boldsymbol{z}\| \) grows. Consequently, the absolute value of the directional derivative satisfies:

\[
\left| \bigl\langle
      \nabla \mathcal{L}_{\text{CE}}(\boldsymbol{z}),
      \;\hat{u}
   \bigr\rangle \right| \to 0 \quad \text{as } \|\boldsymbol{z}\| \to \infty.
\]

This shows that the outward pull from cross-entropy weakens as \( \|\boldsymbol{z}\| \) increases.

\paragraph{Existence of an equilibrium point.}

Since the outward pull from cross-entropy decreases in magnitude as \( \|\boldsymbol{z}\| \) increases, while the inward pull from compression grows linearly, there exists a stable radius \( R \) at which these opposing forces balance:

\[
    \nabla \mathcal{L}_{\text{CE}}(\boldsymbol{z}) + \nabla \mathcal{L}_{\text{2}}(\boldsymbol{z}) = 0.
\]

This equilibrium radius \( R \) prevents the latent representations from collapsing to the origin while still ensuring that they remain confined within a narrow region around \( \boldsymbol{W}_{\overline{y}} \).

\subsection{Combined Geometric Analysis and Volume Confinement}
\label{sec:combined_volume}

We now combine angular and radial arguments to analyze why the latent representations
for each class remain in a thin shell of radius~\(R\) (and why that shell becomes
thinner with increasing \(\gamma\)).  

\begin{lemma}[Volume Confinement via Angular Deviations]
\label{lemma:combined_volume}
Under Assumptions~\ref{assump:separation} and~\ref{assump:gradients}, there exist
constants \(R>0\) and \(\delta R>0\) such that, for sufficiently large \(\gamma\),
the latent representations \(\{\boldsymbol{z}\}\) of each class \(\overline{y}\)
lie in the radial interval \([\,R - \delta R,\,R + \delta R\,]\). Moreover,
\(\delta R = \mathcal{O}(\tfrac{1}{\gamma})\). Consequently, the accessible volume
for each class is bounded by
Here is the expression split into two different lines:

\[
\begin{split}
&\bigl(\text{shell thickness}\bigr) \times \\
\times \bigl(&\text{surface area of the }(d-1)\text{-sphere of radius }R\bigr),
\end{split}
\]

so that
\[
   \text{Volume}(\text{class shell})
   \;\le\;
   C_d\,R^{\,d-1}\,\delta R
   \;\propto\;\frac{R^{\,d-1}}{\gamma},
\]
where \(C_d\) is a constant depending only on dimension~\(d\).  In three
dimensions \((d=3)\), for instance, the shell volume is at most
\[
   4\pi\,R^2\,\delta R
   \;\propto\;\frac{R^2}{\gamma}.
\]
\end{lemma}

\noindent
\textbf{Proof Sketch.}
Consider a latent representation \(\boldsymbol{z}\) belonging to class
\(\overline{y}\) with \(\|\boldsymbol{z}\| = R\).  If \(\boldsymbol{z}\) undergoes
a small angular deviation from its alignment with the class weight vector
\(\boldsymbol{W}_{\overline{y}}\), then to maintain a similar correct‐class logit
\(\boldsymbol{W}_{\overline{y}}^\top \boldsymbol{z}\), it may compensate by
\emph{increasing} \(\|\boldsymbol{z}\|\) slightly (i.e., increasing \(R\)).

Such an increase does not necessarily elevate the total loss: the cross-entropy term
tends to decrease when \(\|\boldsymbol{z}\|\) increases, because the logit magnitude
for class~\(\overline{y}\) becomes larger relative to other classes.  However,
the regularization term \(\mathcal{L}_{\mathrm{H}}(\boldsymbol{z}) = \gamma\|\boldsymbol{z}\|^2\)
imposes an additional penalty:
\[
   \Delta \mathcal{L}_{\mathrm{2}}
   \;=\;
   \gamma\,\bigl((R + \delta R)^2 - R^2\bigr)
   \;\approx\;
   2\,\gamma\,R\,\delta R,
\]
for even a small radial increment \(\delta R\).  In the terminal phase
(Lemma~\ref{lemma:softmax} and Assumption~\ref{assump:separation}),
the cross-entropy gradient is effectively bounded.  Hence, this linear‐in‐\(R\)
compression cost forces \(\delta R\) to shrink with increasing~\(\gamma\).  

More formally, define
\(\delta R(\gamma)\) to be the maximal radial deviation from some
\emph{stable radius} \(R\) such that the loss gradient
remains below a threshold $\epsilon$.  Balancing
the compressive force \(2\,\gamma\,\boldsymbol{z}\) with the bounded outward
pull from cross-entropy yields
\[
   \delta R(\gamma)
   \;\le\;
   \frac{\kappa}{\gamma},
\]
for some \(\kappa>0\).  This implies
\(\lim_{\gamma \to \infty}\delta R(\gamma) = 0\).  Thus, the set of possible
\(\boldsymbol{z}\) for each class collapses to a thin shell of thickness
\(\delta R(\gamma)\) around~\(R_{\overline{y}}\).  Since the \((d-1)\)-dimensional
surface area of a sphere of radius~\(R\) is \(C_d\,R^{\,d-1}\), the total volume
of this shell is at most \(C_d\,R^{\,d-1}\,\delta R(\gamma)\), which in turn is
\(\mathcal{O}\bigl(R^{\,d-1}/\gamma\bigr)\).  \qed

\subsection{Final Collapse within Each Class}
\label{sec:final_collapse_within_class}

\paragraph{Directional alignment.}
By Assumption~\ref{assump:separation}, each training sample \(\boldsymbol{z}\)
satisfies
\(\mathbf{W}_{\overline{y}}^\top\boldsymbol{z} > \mathbf{W}_i^\top\boldsymbol{z}\)
for all \(i\neq \overline{y}\).  Combined with Lemma~\ref{lemma:softmax},
this implies \(p_{\overline{y}}(\boldsymbol{z}) \approx 1\).  Substantial
angular deviations from the direction of \(\mathbf{W}_{\overline{y}}\)
would lower \(\mathbf{W}_{\overline{y}}^\top \boldsymbol{z}\) and thus
raise the cross-entropy term.  Consequently, in the terminal phase,
\(\boldsymbol{z}\) remains close to the direction of \(\mathbf{W}_{\overline{y}}\).

\paragraph{Radial collapse.}
Simultaneously, the penalty \(\gamma\,\|\boldsymbol{z}\|^2\) discourages large norms
\(\|\boldsymbol{z}\|\).  Maintaining strict separation while preserving high‐confidence
logits restricts \(\boldsymbol{z}\) to a narrow band around
\(\mathbf{W}_{\overline{y}}\).  From Lemma~\ref{lemma:combined_volume} (and its
limit argument), we have
\(\lim_{\gamma\to\infty}\delta R(\gamma)=0\).  Hence, as \(\gamma\) grows,
class‐\(\overline{y}\) samples must \emph{collapse} in radius, converging to
a stable radius \(R\) with diminishing fluctuations.

\begin{theorem}[Latent Point Collapse]
\label{thm:collapse}
Under Assumptions~\ref{assump:separation} and~\ref{assump:gradients}, there exists a radius \(R_{\overline{y}}\) such that, for sufficiently large
\(\gamma\), every latent representation \(\boldsymbol{z}\) 
lies in a thin shell of thickness \(\delta R(\gamma)\) around
\(\|\boldsymbol{z}\|=R\).  Moreover,
\(\lim_{\gamma\to\infty} \delta R(\gamma)=0\).  Consequently, each class collapses
onto a unique point (up to vanishingly small fluctuations) in \(\mathbb{R}^d\).
Formally, for any \(\epsilon>0\), there exists \(\Gamma>0\) such that for all
\(\gamma>\Gamma\),
\[
  \max_{\boldsymbol{z}\in\mathcal{D}_{\overline{y}}}
  \|\boldsymbol{z} - \boldsymbol{z}_{\overline{y}}^*\|
  \;\le\;\epsilon,
\]
where \(\boldsymbol{z}_{\overline{y}}^*\) is the limiting latent point for class
\(\overline{y}\) and \(\mathcal{D}_{\overline{y}}\) denotes the set of training
points of that class.
\end{theorem}

\noindent
\textbf{Interpretation.}
This result formalizes \emph{latent point collapse}: all samples of a given class
\(\overline{y}\) converge to (nearly) the same vector in \(\mathbb{R}^d\),
reflecting both a strict directional alignment with \(\mathbf{W}_{\overline{y}}\)
and a converging norm dictated by the balance of cross‐entropy and compression.
As \(\gamma\) grows, fluctuations around that limiting point vanish, leading to
\(\delta R(\gamma)\to 0\).  The class‐wise separation (Assumption~\ref{assump:separation})
prevents \(\boldsymbol{z}\) from collapsing to the origin entirely, instead favoring
a stable radius that satisfies the margin constraints on other classes.

\section{Information Bottleneck in Deterministic DNN Classifiers}
\label{App:IB}
The IB objective can be formulated as an optimization problem \cite{tishby2000informationbottleneckmethod}, aiming to maximize the following function:

\begin{equation*}
\label{eq:loss_ib_app}
\mathcal{L}_{IB} = I(\mathbf{z}; y) - \beta I(\mathbf{z}; \mathbf{x}),
\end{equation*}

where \( I(\mathbf{z}; y) \) denotes the mutual information between the latent representation \( \mathbf{z} \) and the labels \( y \), while \( I(\mathbf{z}; \mathbf{x}) \) represents the mutual information between \( \mathbf{z} \) and the input data \( \mathbf{x} \). The parameter \( \beta \) controls the trade-off between compression and predictive accuracy.
Our goal is to maximize the mutual information between the latent representations and the labels, \( I(\mathbf{z}; y) \). This mutual information can be expressed in terms of entropy:

\begin{equation*}
I(\mathbf{z}; y) = H(y) - H(y|\mathbf{z}),
\end{equation*}

where \( H(y) \) is the entropy of the labels and \( H(y|\mathbf{z}) \) is the conditional entropy of the labels given the latent representations. Since \( H(y) \) is constant with respect to the model parameters (as it depends solely on the distribution of the labels), maximizing \( I(\mathbf{z}; y) \) is equivalent to minimizing the conditional entropy \( H(y|\mathbf{z}) \):

\begin{equation*}
\max I(\mathbf{z}; y) \quad \Leftrightarrow \quad \min H(y|\mathbf{z}).
\end{equation*}

The conditional entropy \( H(y|\mathbf{z}) \) can be estimated empirically using the dataset. Assuming that the data points \( (\mathbf{x}^{(n)}, y^{(n)}) \) are sampled from the joint distribution \( p(\mathbf{x}, y) \) and that \( \mathbf{z}^{(n)} = f(\mathbf{x}^{(n)}) \), we approximate \( H(y|\mathbf{z}) \) as:

\begin{equation*}
H(y|\mathbf{z}) \approx -\frac{1}{N} \sum_{n=1}^N \sum_{k=1}^K p(y_k|\mathbf{z}^{(n)}) \log p(y_k|\mathbf{z}^{(n)}),
\end{equation*}

where \( K \) is the number of classes and \( p(y_k|\mathbf{z}^{(n)}) \) is the probability of label \( y_k \) given latent representation \( \mathbf{z}^{(n)} \). In practice, since we have the true labels \( y^{(n)} \), this simplifies to:

\begin{equation*}
H(y|\mathbf{z}) \approx -\frac{1}{N} \sum_{n=1}^N \log p(y^{(n)}|\mathbf{z}^{(n)}).
\end{equation*}

This expression corresponds to the cross-entropy loss commonly used in training classifiers. In a DNN classifier, the probability \( p(y|\mathbf{z}) \) is modeled using the softmax function applied to the output logits:

\begin{equation*}
p(y_k|\mathbf{z}) = \frac{\exp\left( (\mathbf{W}\mathbf{z} + \mathbf{b})_k \right)}{\sum_{i=1}^K \exp\left( (\mathbf{W}\mathbf{z} + \mathbf{b})_i \right)},
\end{equation*}

where \( \mathbf{W} \) and \( \mathbf{b} \) are the weights and biases of the final layer, and \( (\mathbf{W}\mathbf{z} + \mathbf{b})_k \) denotes the logit corresponding to class \( y_k \). By minimizing \( H(y|\mathbf{z}) \), we encourage the model to produce latent representations that are informative about the labels, aligning with the objective of accurate classification.

The second term in the IB objective, \( I(\mathbf{z}; \mathbf{x}) \), quantifies the mutual information between the latent representations and the inputs. To achieve compression, we aim to minimize this term. Expressing \( I(\mathbf{z}; \mathbf{x}) \) in terms of entropy:

\begin{equation*}
I(\mathbf{z}; \mathbf{x}) = H(\mathbf{z}) - H(\mathbf{z}|\mathbf{x}).
\end{equation*}

In the case of deterministic mappings where \( \mathbf{z} = f(\mathbf{x}) \), the differntial conditional entropy \( H(\mathbf{z}|\mathbf{x}) \) is ill-defined, therefore we focus solely on minimizing  $H(\mathbf{z})$ as explained in the InfoMax seminal paper \cite{infomax}.

\begin{equation*}
\min I(\mathbf{z}; \mathbf{x})  \quad \Leftrightarrow \quad  \min H(\mathbf{z}).
\end{equation*}

\section{Training and Architecture Details.} 
\label{App:training_details}

\begin{table*}[t]
\label{table:summary_architectures}
\caption{Summary of the features implemented in all architectures used in our ablation study. \textit{Lin. Pen} refers to the inclusion or exclusion of a linear penultimate layer. \textit{Nodes Add. Layer} feature indicates the presence of an additional layer between the backbone and the classification layer. If this layer is present, its dimensionality is categorized as one of three possible values: wide, intermediate, or narrow. The exact dimensionality for these categories is a hyperparameter that varies across different datasets. 
\textit{Loss} indicates the type of loss function utilized during training.
}
\label{table:architectures}
\vskip 0.15in
\begin{center}
\begin{small}
\begin{sc}
\begin{tabular}{lccl}
\toprule
Model   & Lin. Pen. & Nodes Add. Layer  & Loss\\
\midrule
LPC             & \cmark  & Intermediate & $CE  +L_2$\\
LPC-Wide         & \cmark &  Wide & $CE  +L_2$\\
LPC-Narrow       & \cmark &  Narrow  & $CE +L_2$\\
LPC-SCL       & \cmark &  Intermediate & $CE  +L_2 +SCL$ \\
LPC-NoPen       & \xmark &  \xmark & $CE  +L_2$\\
\cmidrule(lr){1-4}
LinPen         & \cmark &  Intermediate  & $CE$\\
NonlinPen      & \xmark &  Intermediate & $CE$ \\
SCL     & \xmark &  \xmark  & $CE + SCL$\\
Arcface     & \xmark &  \xmark  & $ArcFace$\\
CosFace     & \xmark &  \xmark & $CosFace$ \\
NoPen          & \xmark &  \xmark & $CE$ \\
\bottomrule
\end{tabular}
\end{sc}
\end{small}
\end{center}
\vskip -0.1in
\end{table*}

To generate the latent representation $\boldsymbol{h}(\boldsymbol{x})$, we employed two distinct ResNet~\cite{resnet} backbone architectures of increasing complexity for two different datasets. Specifically, a ResNet-18 architecture was used for the CIFAR-10~\cite{cifar} dataset, while a ResNet-50 architecture was used for the CIFAR-100~\cite{cifar} dataset. Both ResNet architectures included batch normalization and used the Swish activation function~\cite{swish} in all non-linear layers.

The \textsc{LPC}, \textsc{LPC-SCL}, \textsc{LinPen}, and \textsc{NonlinPen} architectures included an additional fully connected penultimate layer of 64 nodes for CIFAR-10 and 128 nodes for CIFAR-100. The \textsc{LPC-Wide} architecture used 128 nodes for CIFAR-10 and 256 nodes for CIFAR-100 in the penultimate layer, whereas the \textsc{LPC-Narrow} architecture employed 16 nodes for CIFAR-10 and 32 nodes for CIFAR-100. Meanwhile, the \textsc{LPC-NoPen}, \textsc{NoPen}, \textsc{SCL}, \textsc{ArcFace}, and \textsc{CosFace} architectures did not include any additional penultimate layer. A summary of the different architectures is provided in Table~\ref{table:architectures}.

For training, we used the AdamW~\cite{kingma2017adam,adamw} optimizer with default PyTorch settings and a weight decay of $0.5 \times 10^{-4}$. Data augmentation consisted of random horizontal flips and random cropping (with padding $=4$). Each experiment was repeated across different learning rates, chosen from a geometric sequence starting at $10^{-4}$ and increasing by a factor of 2 for a total of four distinct values. The best session on the test set (in the final epoch) was selected. The bias $b$ of the last linear classifier was set to zero as in~\cite{Deng_2022}. Starting from epoch 250, learning rates were halved every 50 epochs, except for the last linear classifier, which was unaffected by this schedule.

All training sessions ran for 800 epochs, recording performance metrics every 50 epochs. We employed extended simulations so that most training occurs in the terminal phase of training (TPT), defined as the period after the network achieves 99.9\% training accuracy~\cite{neural_collapse}.

For architectures using an $L_2$ loss, the coefficient $\gamma$ started at a small value ($10^{-2}$) and was multiplied by $\gamma_{step}$ each epoch until it reached $\gamma_{max}$. After reaching $\gamma_{max}$, it remained constant. For both CIFAR-10 and CIFAR-100, we set $\gamma_{step} = 1.05$ and $\gamma_{max} = 10^6$.

The SCL loss function~\cite{khosla2021supervisedcontrastivelearning} is given by
\[
\mathcal{L}^{\text{SCL}} 
= -\frac{1}{N} 
  \sum_{i=1}^{N} 
    \frac{1}{|P(i)|} 
    \sum_{p \in P(i)} 
      \log \frac{\exp(s_{i,p} / \tau)}
                {\sum_{j \neq i} \exp(s_{i,j} / \tau)},
\]
where $N$ is the batch size, $P(i)$ is the set of positive samples for instance $i$, $|P(i)|$ is its cardinality, $s_{i,j}$ is the cosine similarity between samples $i$ and $j$, and $\tau$ is the temperature (set to a low value $0.05$ to increase margins in latent space). Since minimizing only $\mathcal{L}^{\text{SCL}}$ is insufficient for classification, we also minimized the cross-entropy loss in parallel.

For ArcFace and CosFace, we followed the margin-based strategies in~\cite{Deng_2022,wang2018cosfacelargemargincosine}. In both cases, a scaled cosine similarity is modified by a margin before applying softmax and computing the cross-entropy. ArcFace applies an angular margin $m$ (increasing from $0.1$ to $0.5$), while CosFace subtracts a margin $m$ (increasing from $0.05$ to $0.25$) directly from $\cos \theta$. In both cases, the scale factor $s$ is gradually increased from 16 to 64 during training.

\[
\mathcal{L}^{\text{ArcFace}} = \frac{1}{N} \sum_{i=1}^{N} 
\text{CE}\Bigl(\text{Softmax}\bigl[s \cdot (\cos(\theta_i + m))\bigr], y_i\Bigr),
\]
\[
\mathcal{L}^{\text{CosFace}} = \frac{1}{N} \sum_{i=1}^{N}
\text{CE}\Bigl(\text{Softmax}\bigl[s \cdot (\cos(\theta_i) - m)\bigr], y_i\Bigr).
\]

Finally, for each architecture and dataset, five independent experiments were conducted using different random initializations. Reported results in all plots are the mean and standard deviation of those trials.

\section{Binarity Hypothesis} 
\label{App:binarity_hypothesis}
\begin{figure*}[t]
\vskip 0.1in
\begin{center}
    \centering
    \includegraphics[width=\textwidth]{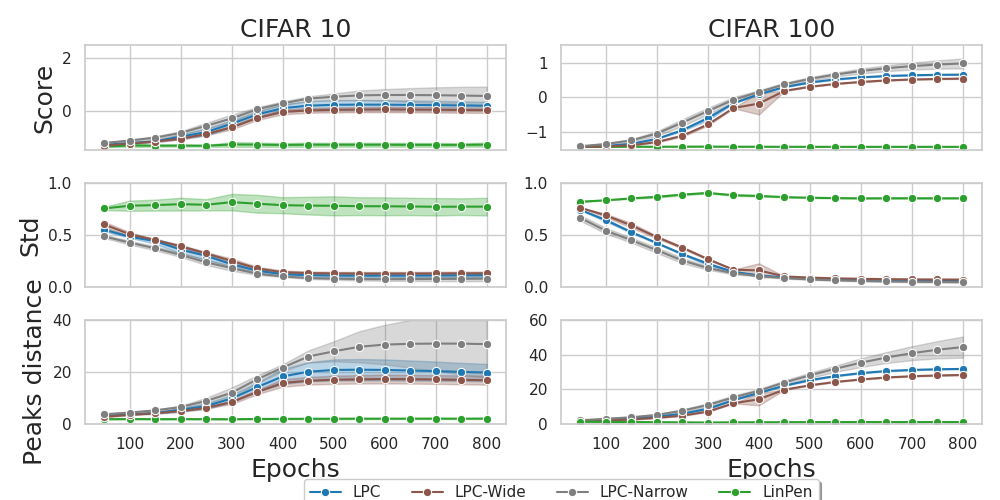}
    \vskip 0.1in
    \caption{Log-likelihood scores, standard deviations and weighted peak distance of bimodal Gaussian mixture models fitted on each dimension of the penultimate layer using all values of the training set. From top to bottom, the quantities described as $\overline{\ell}$, $\overline{\sigma}$, and $\overline{\mu}$ in Eqs. \ref{Eq:score_likelihood}, \ref{Eq:score_stds}, \ref{Eq:score_peaks} are computed for the \textsc{LPC}, \textsc{LPC-Wide}, \textsc{LPC-Narrow}, \textsc{LinPen} architectures. We can see that the latent representations in the penultimate layers featuring $L_2$ loss are well represented by two Gaussians with increasingly small standard deviations, while this is not observed for the \textsc{LinPen}  architectures.}
    \label{Fig:score} 
\hfill
\end{center}
\vskip -0.1in
\end{figure*}
\begin{figure}[t]
\vskip 0.1in
\begin{center}

\end{center}
\vskip -0.1in
\end{figure}

\begin{table*}[t]
\caption{Score $\overline{\ell}$ and relative distance of the two distances $\overline{\mu}$ over all penultimate nodes in the training set across all experiments at the last epoch. Average and min values are shown. The coefficient of variation measures the standard deviation of the norm of latent representations normalized by the mean.}
\label{table:bin_hyp}
\vskip 0.15in
\begin{center}
\begin{small}
\begin{sc}
\begin{tabular}{lccccc}
\toprule
\multicolumn{6}{c}{Dataset: CIFAR10} \\
\cmidrule(lr){1-6}
Model & Score & Min Score & Peaks Dist & Min Peaks Dist & Coeff. of Var. \\
\midrule
LPC & 0.181 $\pm$ 0.154 & -0.331 & 19.810 $\pm$ 3.329 & 11.468 & 0.204 \\
LPC-Wide & 0.011 $\pm$ 0.106 & -0.540 & 16.897 $\pm$ 1.699 & 9.409 & 0.232 \\
LPC-Narrow & 0.559 $\pm$ 0.358 & 0.139 & 30.726 $\pm$ 12.860 & 18.727 & 0.174 \\
LPC-SCL & -0.057 $\pm$ 0.116 & -0.558 & 15.839 $\pm$ 1.597 & 9.163 & 0.245 \\
LinPen & -1.307 $\pm$ 0.081 & -1.421 & 2.174 $\pm$ 0.154 & 0.439 & 0.953 \\
\addlinespace
\multicolumn{6}{c}{Dataset: CIFAR100} \\
\cmidrule(lr){1-6}
Model & Score & Min Score & Peaks Dist & Min Peaks Dist & Coeff. of Var. \\
\midrule
LPC & 0.658 $\pm$ 0.005 & 0.476 & 31.832 $\pm$ 0.153 & 26.380 & 0.077 \\
LPC-Wide & 0.542 $\pm$ 0.007 & 0.308 & 28.336 $\pm$ 0.190 & 22.389 & 0.083 \\
LPC-Narrow & 0.980 $\pm$ 0.143 & 0.756 & 44.379 $\pm$ 6.085 & 35.272 & 0.060 \\
LPC-SCL & 0.385 $\pm$ 0.121 & 0.185 & 24.383 $\pm$ 3.028 & 19.834 & 0.094 \\
LinPen & -1.420 $\pm$ 0.000 & -1.423 & 1.252 $\pm$ 0.011 & 0.463 & 0.800 \\
\addlinespace
\midrule
\bottomrule
\end{tabular}
\end{sc}
\end{small}
\end{center}
\vskip -0.1in
\end{table*}

Our assumption is that each dimension on the penultimate latent representation can assume approximately only one of two values.
In order to verify this assumption, we fit a Gaussian mixture model (GMM) with 2 modes on each set of latent representations
$z_i \sim \mathcal{N}\left(\mu^{(1,2)}_{ i}; {\sigma_{i}^{(1,2)}}^2\right).$
For each dimension $i$, we build a histogram with the values of all latent representations of the training set. 
We then fit a bimodal GMM model on this histogram.
Assuming that $P$ is the dimensionality of the latent representation and the dataset contains $N$ datapoints, the following quantities are collected:  
The average log-likelihood score
\begin{equation}
    \overline{\ell} = \frac{1}{NP}\sum_{n=0}^{N-1}\sum_{i=0}^{P-1}\log\mathcal{N}\left(z^{(n)}_i \middle|\mu^{(1,2)}_{ i};{\sigma_{i}^{(1,2)}}^2\right);   
\label{Eq:score_likelihood}
\end{equation}
the average standard deviation of the two posterior distributions
\begin{equation}
       \overline{\sigma}=\frac{1}{P}\sum_{i=0}^{P}\left( \frac{\sigma_i^{(1)}+\sigma_i^{(2)}}{2} \right);
    \label{Eq:score_stds} 
\end{equation}
and the mean relative distance of the two peaks reweighted with the standard deviation
\begin{equation}
  \overline{\mu} = \frac{1}{P}\sum_{i=0}^{P} \frac{\left\|\mu_i^{(2)}-\mu_i^{(1)}\right\|}{\left(\sigma_i^{(1)}+\sigma_i^{(2)}\right)/2}.     
\label{Eq:score_peaks}  
\end{equation}

These values are plotted in Fig. \ref{Fig:score}, showing that during training, the log-likelihood score increases while the standard deviation decreases for all \textsc{LPC} models. Additionally, the relative distance between the two modes of the GMM models increases. These three metrics indicate that during training, all latent representations collapse into two distinct points, forming two clearly separated clusters.
This observation supports our binarity hypothesis, which states that each latent representation can assume only one of two possible values. The same analysis was performed for the \textsc{LinPen} architecture, which also features a linear layer before classification. However, in this architecture, the binarity hypothesis does not hold.

To validate our claim that binary encoding manifests in each node of the penultimate layer of \textsc{LPC} models, we present averages in Table \ref{table:bin_hyp}. These averages are computed across all nodes in the penultimate layer from all experiments. The table shows the average and minimum values for the GMM fitting score and the weighted relative distance between the peaks across all nodes and experiments.
For all architectures—defined as those with a linear layer, latent compression, and a dimensional bottleneck—the binarity hypothesis holds true for all dimensions, even in cases with the lowest recorded scores.
The table also includes the coefficient of variation for the absolute values of the latent representations evaluated for all dimensions and samples. 
The low values of the coefficient of variation in the \textsc{LPC} architectures indicate that in each node they can assume approximately only one of two possible values.
This observation supports our claim that the class means of the latent representations collapse onto the vertices of a hypercube.

\section{Neural collapse}
\label{App:neural_collapse}
\begin{figure*}[t]
\vskip 0.1in
\begin{center}
\centerline{\includegraphics[width=5.3in]{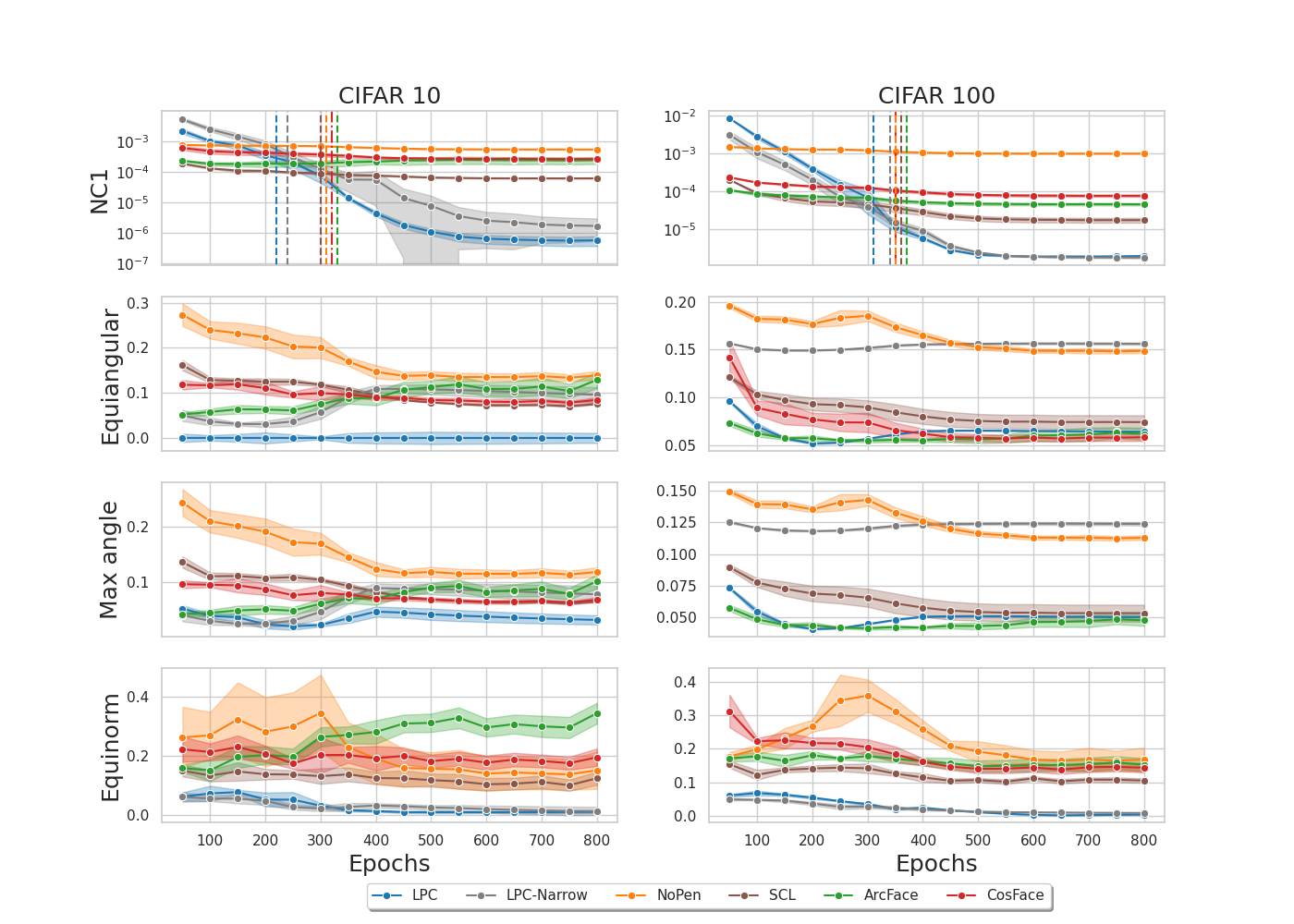}}
\vskip 0.1in
\caption{Metrics used to evaluate convergence towards Neural Collapse (NC). In the upper figure, we examine a renormalized version of the NC1 property. This normalization process is conducted based on the number of nodes in the penultimate layer to ensure a fair comparison across models with varying dimensions of the penultimate layer. The dashed lines are drawn at the average epoch when training reaches convergence, that demonstrates that most of the training was performed in the TPT.
Below, we present metrics demonstrating convergence to an ETFS, utilizing the same parameters as those outlined in \cite{neural_collapse}.}
\label{Fig:nc_metrics} 
\end{center}
\vskip -0.1in
\end{figure*}

In this appendix, we present all metrics related to NC as defined in in \cite{neural_collapse}. 
The entire NC phenomenon can be summarized into four distinct components: (1) the variability of samples within the same class diminishes as they converge to the class mean (NC1); (2) the class means in the penultimate layer tend towards an ETFS (NC2); (3) the last layer classifier weights align with the ETFS in their dual space (NC3); and (4) classification can effectively be reduced to selecting the closest class mean (NC4).

The first property of interest is NC1, which asserts that the variability of samples within the same class decreases in the terminal phase of training. This property is characterized by the equation $\text{Tr}\left(\boldsymbol{\Sigma}_W\boldsymbol{\Sigma_B}^\dagger/K\right)$, where $\Sigma_W$ is defined as 

\begin{equation}
\Sigma_{W}=\frac{1}{NP}\sum_{i=0}^{N-1}\sum_{p=0}^{P-1}\left(\boldsymbol{z}^{(i,p)}-\boldsymbol{\mu}^{(p)} \right)\left(\boldsymbol{z}^{(i,p)}-\boldsymbol{\mu}^{(p)} \right)^\top
\end{equation}    

where $\boldsymbol{z}^{(i,p)}$ is the $i$-th latent representation with label $p$ and, $\boldsymbol{\mu}^{(p)}$ is the mean of all representation with label $p$; and $\Sigma_B$ is defined as:

\begin{equation}
\Sigma_{B}=\frac{1}{P}\sum_{p=0}^{P-1}\left(\boldsymbol{\mu}^{(p)}-\boldsymbol{\mu}_G \right)\left(\boldsymbol{\mu}^{(p)}-\boldsymbol{\mu}_G \right)^\top.
    \label{eq:sigma_b}
\end{equation}
 
The trace operation sums over all diagonal elements, the dimensionality of which is equal to that of the penultimate layer, $P$. 
Given the use of different architectures with varying numbers of nodes in the penultimate layers in our study, we examine a renormalized version of this quantity, $\text{Tr}\left(\boldsymbol{\Sigma}_W\boldsymbol{\Sigma_B}^\dagger/K/P\right)$.

In Fig \ref{Fig:nc_metrics}, the top image presents this value, showing that it is at least an order of magnitude lower in the \textsc{LPC} architectures compared to the baseline architecture. 
We also note that the other regularization techniques SCL, ArcFace, and CosFace, provide better convergence to NC with respect to the baseline, but improvements remain lower with respect LPC models.

The other three images below demonstrate the convergence of class means toward an ETFS, also known as the NC2 property. These images show that all values reach a plateau in the terminal phase, indicating convergence to their optimal values. It is evident that the \textsc{LPC-Narrow} architecture, which uses a smaller-dimensional embedding in the penultimate layer, tends to exhibit higher values for the angular measures (maximum angle and equiangularity) compared to the baseline. This is because, geometrically, it is more challenging for the network to construct an ETFS using the vertices of a hypercube in a low-dimensional space. 

By observing the metrics in Fig. \ref{Fig:nc_metrics}, we conclude that while regularization techniques accelerate convergence to NC, the best convergence is achieved with LPC. We also note that the dashed lines represent the average epoch at which the network reached convergence, showing that most of the training occurred after convergence, in the TPT. All metrics have reached plateaus, demonstrating that the phenomenon of NC is fully realized. Thus, the additional benefits of LPC documented in this paper are in addition to those typically associated with NC.

\end{document}